\documentclass[11pt,a4paper]{article}

\usepackage[T1]{fontenc}
\usepackage[utf8]{inputenc}
\usepackage{a4wide}
\usepackage{newtxtext}
\usepackage{graphicx}
\usepackage{amsmath}
\usepackage{newtxmath}
\usepackage{microtype}
\usepackage[round,authoryear]{natbib}
\usepackage[
  colorlinks=true,
  linkcolor=blue,
  citecolor=blue,
  urlcolor=blue
]{hyperref}

\title{Shared Organizational Memory for Enterprise Coding Agents: System Design and Deployment Snapshot}

\author{
  \textbf{Harsh Rao Dhanyamraju}\\
  \emph{Data Scientist}\\
  SAP SE
  \and
  \textbf{Leonidas Raghav}\\
  \emph{Machine Learning Engineer}\\
  SAP SE
}

\date{July 2026\\Preprint}

\begin{document}
\maketitle

\begin{abstract}
Enterprise coding agents rely on tools and retrieval, yet enterprise knowledge often remains outside public training data and formal documentation: internal DSLs, proprietary platforms, local conventions, recent fixes, and tacit workflows. Existing knowledge interfaces expose stored resources but still depend on agents recognizing and explicitly recording lessons worth reusing, disconnecting capture from the coding workflow and leaving development experience repeatedly rediscovered. We report an ongoing production deployment of a shared organizational memory system that makes capture a platform-level part of coding work: it collects task-adjacent experience with contributor approval, curates it into reusable question--answer memories, gates obvious security and privacy risks, and retrieves memories for future agents. This short paper describes the deployed lifecycle and an operational snapshot; effects on retrieval and coding tasks remain under evaluation.
\end{abstract}

% The body below consolidates intro.md and Sections 2--6. The three files in
% 1-intro/ are component drafts already incorporated into intro.md, and
% realm-2026-section-plan.md is planning metadata rather than manuscript text.

\section{Introduction}

Enterprise software relies on domain-specific languages (DSLs) and internal languages to encode business, operational, and platform knowledge. \citet{mernik2005} and \citet{vandeursen2000} show how DSLs expose domain concepts, constraints, and idioms directly in the programming surface, while \citet{hermans2009} demonstrates how they make organizational policies and workflows programmable and repeatable. Examples and programming models include SAP ABAP \citep{sap-abap}, the SAP Cloud Application Programming Model (CAP) \citep{sap-cap}, Salesforce SOQL \citep{salesforce-soql}, Terraform/HCL \citep{terraform-language}, Kusto Query Language \citep{kql}, Bazel Starlark \citep{starlark}, Atlassian JQL \citep{jql}, and Shopify Liquid \citep{liquid}. This creates a difficult setting for coding assistants: even public DSLs present version-specific behavior, edge cases, and organization-specific conventions that may be poorly documented. Internal DSLs, including proprietary configuration, policy, schema, build, and deployment languages, may be absent from public training data altogether. In these cases, the model has no reliable prior and must learn from the repository, surrounding tools, and outcomes of earlier development work.

Recent coding-agent research shows that performance depends heavily on access to the development environment. \citet{jimenez2024swebench} introduced issue-resolution tasks drawn from real repositories, while \citet{yang2024sweagent} and \citet{zhang2024autocoderover} demonstrated the importance of agent-computer interfaces, repository navigation, program analysis, and context retrieval. Practical harnesses such as Claude Code \citep{claude-code}, OpenAI Codex \citep{codex}, Google Antigravity \citep{antigravity}, and Cursor \citep{cursor} reflect the same need for structured access to files, tools, tests, and developer feedback. Tool and knowledge interfaces extend this context: the Model Context Protocol \citep{mcp-spec} connects agents to external data through reference servers \citep{mcp-servers}, Playwright MCP \citep{playwright-mcp}, and Context7 \citep{context7}, while Claude Code Skills \citep{claude-code-skills} and Anthropic Agent Skills \citep{anthropic-agent-skills} package reusable instructions and resources. Yet access is not knowledge: these interfaces can expose files, APIs, and documentation without capturing organization-specific conventions, recent fixes, internal DSL semantics, or tacit workflows. Documentation may also be incomplete or stale, and private behavior may become clear only through debugging, tests, failures, and fixes.

Agentic memory provides enterprise coding agents with external, persistent state---such as notes, prior fixes, curated skills, and repository observations---rather than modifying model weights or hidden state. Surveys characterize memory as a durable layer distinct from transient context \citep{jiang2026anatomy,hu2025memory}, while SWE-MeM \citep{gao2026swemem}, MemDocAgent \citep{bae2026memdoc}, and SWE-EVO \citep{le2025sweevo} show the value of compressing, retrieving, and reusing prior development state. Anthropic's harness guidance \citep{anthropic-harnesses} similarly uses commits and progress artifacts across sessions. For enterprise agents, memory allows private development experience to accumulate instead of being repeatedly rediscovered.

Mozilla AI's CQ \citep{mozilla-cq} demonstrates how agents can share structured Knowledge Units. Its architecture \citep{cq-architecture} supports querying, proposing, confirming, flagging, and promoting knowledge for reuse. The enterprise gap, however, appears at both ends of this lifecycle. During capture, its architecture \citep{cq-architecture} and quickstart \citep{cq-quickstart} require agents to invoke tools such as query, propose, or reflect. Because an agent's immediate objective is completing a coding task, useful lessons may never be contributed if the agent does not recognize their future value. Human confirmation and review also become difficult to scale across large organizations. During consumption, CQ acts as a general shared knowledge layer, whereas enterprise coding agents need relevant memories delivered when they are navigating, editing, testing, or debugging. CQ therefore enables knowledge sharing once contribution occurs, but does not make capture automatic or retrieval native to the coding workflow.

Nevertheless, a central gap remains: existing approaches expose tools and shared knowledge interfaces but do not make the capture and reuse of private development experience a first-class part of the coding workflow. We report the current architecture and production deployment of an organizational memory layer that automatically initiates the capture of task-adjacent knowledge, subject to contributor approval; curates accepted contributions under provenance and policy controls; and retrieves relevant memories for future agents working across repositories, DSLs, and development tools.

\section{System Architecture}

\begin{figure}[htbp]
  \centering
  \includegraphics[
    alt={Three-stage organizational memory architecture. A contributor client collects approved local knowledge changes; a curation pipeline enriches, tags, security-scans, deduplicates, scores, and stores memories; and a consumer client retrieves curated memories through MCP, REST, and a user interface.},
    width=0.68\textwidth
  ]{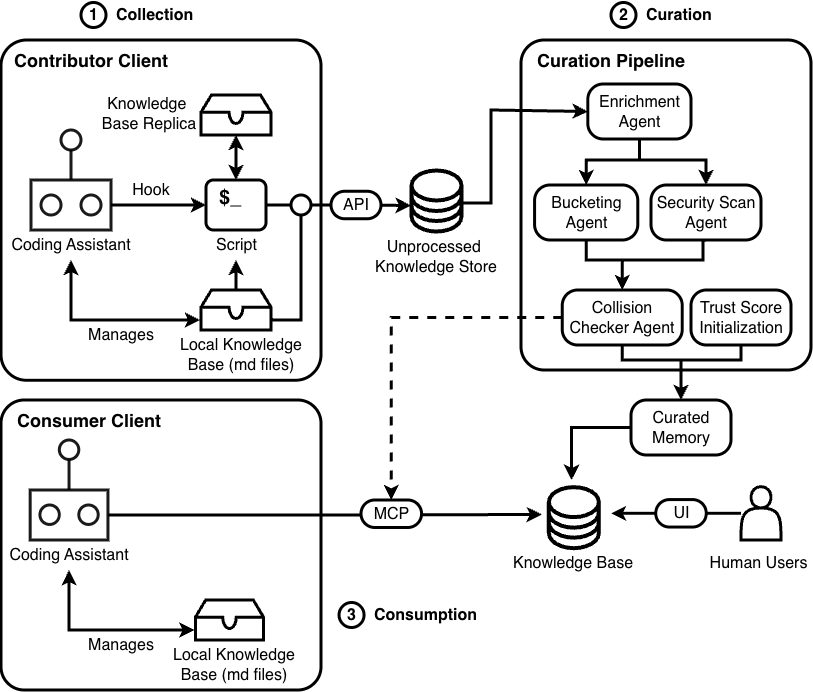}
  \caption{Organizational memory architecture across (1) collection, (2) curation, and (3) consumption.}
  \label{fig:memory-architecture}
\end{figure}

Figure~\ref{fig:memory-architecture} summarizes the lifecycle in three numbered stages: (1) \emph{Collection} in the Contributor Client captures changes to local knowledge, (2) \emph{Curation} in the Curation Pipeline converts approved changes into governed Curated Memory, and (3) \emph{Consumption} in the Consumer Client retrieves that memory for later coding tasks. This separation keeps the stages independently controllable while connecting them through explicit API, MCP, and UI boundaries.

\subsection{Contributor Client}

The Contributor Client turns real changes to local memory rules into candidate knowledge for curation. Setup is enabled through an internal CLI that configures a project to work with the Coding Assistant. The assistant manages a project-scoped Local Knowledge Base of Markdown files, while the client maintains a Knowledge Base Replica as the baseline for detecting changes.

Once enabled, the Coding Assistant is given two Hook events, \texttt{PostToolUse} and \texttt{Stop}, following the Claude Code hooks mechanism \citep{claude-hooks}. \texttt{PostToolUse} runs after file-writing actions such as \texttt{Write}, \texttt{Edit}, and \texttt{MultiEdit}, and marks the memory state as dirty only when a knowledge-base file changes. \texttt{Stop} runs when the assistant finishes its task and invokes the Script.

\begin{figure}[htbp]
  \centering
  \includegraphics[
    alt={Automatic contribution flow. A coding assistant updates the local knowledge base; hooks mark it dirty and automatically run the contribution script at task completion; the script compares it with a replica and builds a diff; the contributor approves or declines; approved diffs are sent through the contribution API to the unprocessed knowledge store.},
    width=0.96\textwidth
  ]{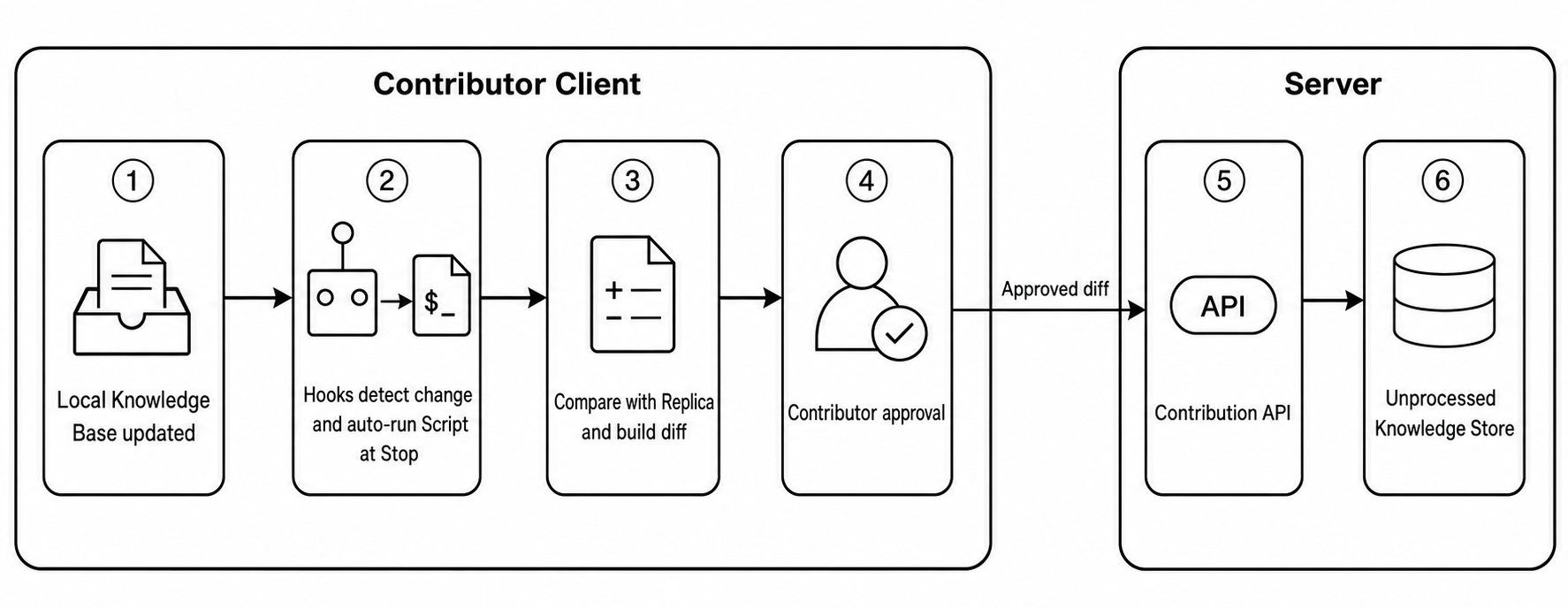}
  \caption{Automatic contribution flow. Hook events initiate comparison and diff construction; transmission to the server occurs only after contributor approval.}
  \label{fig:automatic-contribution-flow}
\end{figure}

Figure~\ref{fig:automatic-contribution-flow} expands region 1 of Figure~\ref{fig:memory-architecture}. The Script compares the Local Knowledge Base against the Knowledge Base Replica, identifies changed regions, prepares the diff, and asks the user for permission before sending an approved candidate through the API. Unlike CQ, contribution is automatically initiated; the user only has to approve it. After a successful contribution, the Script refreshes the replica so that the next comparison uses a stable project-local baseline.

For each changed knowledge-base file, the Script uses Diff Match Patch to construct a stable textual representation of the edit \citep{diff-match-patch}. It also adds lightweight context, including five lines above and below each changed region and the surrounding heading hierarchy extracted with regular expressions. This gives the Curation Pipeline enough local context without asking the Coding Assistant to explain or package its own edits.

This follows the fly-on-the-wall design philosophy: observe task-adjacent memory edits, preserve useful context, ask for approval, and otherwise avoid interfering with the agent's directed work. Earlier versions considered cloud-side diffing, but we ruled it out because it would require either storing each contributor's raw local memories or repeatedly sending enough local state to reconstruct changes.

\subsection{Curation Pipeline}

The Curation Pipeline, shown in region 2 of Figure~\ref{fig:memory-architecture}, turns approved knowledge-base edits into durable, retrievable knowledge. The API places each candidate in the Unprocessed Knowledge Store, from which the pipeline reads the learning and normalizes it into a change record with contributor, team, source category, structured hunks, and local edit context.

The Enrichment Agent converts this record into canonical memories. It first filters hunks with insufficient context, then turns surviving hunks into zero or more self-contained question--answer memories with a developer-facing question, direct answer, copied working examples, and caveats when present. It can merge related hunks, split dense hunks, or output nothing when no reusable knowledge exists.

The enriched output branches to the Bucketing Agent and the deployed Security Scan Agent. The Bucketing Agent assigns tool/platform and knowledge-type tags from the registry. When no existing tag fits a candidate, it can create a new registry tag, subject to a separate tag-validation gate. The Security Scan Agent provides lightweight, deterministic in-repository checks corresponding to OSV-Scanner and GuardDog for dependency and supply-chain risks \citep{osv-scanner,guarddog}, Gitleaks for exposed secrets \citep{gitleaks}, Bandit and Semgrep MCP for unsafe code patterns \citep{bandit,semgrep-mcp}, and Presidio for personal data \citep{presidio}; these are not pinned executions of the external binaries. Its decision contract permits persistence unchanged, persistence of a rewritten memory, or dropping the candidate; it has no quarantine state.

As a functional sanity check of the deployed gate, we evaluated 56 synthetic cases covering dependency risks, secrets, unsafe code, personal or internal information, unsafe workflows, and benign controls. Across three identical runs (168 executions), \textbf{135 (80.4\%)} outcomes were appropriate, \textbf{32 (19.0\%)} were cautious but non-blocking, \textbf{1 (0.6\%)} was overcautious, and \textbf{0 (0.0\%)} were unsafe acceptances (Appendix~\ref{app:gate-eval}).

The Collision Checker Agent searches existing memories using tag-filtered keyword and vector similarity over questions and answers. It passes candidates with no close match; otherwise, it accepts, clarifies, or rejects them as duplicate or low-value and may link retained memories to prerequisites, follow-ups, or alternatives.

Trust Score Initialization derives an initial reliability signal from examples, caveats, and recency and creates agent and human feedback counters. After security and collision checks, the pipeline embeds question and answer separately, persists the Curated Memory in the shared Knowledge Base, initializes feedback tracking, and marks the source learning processed.

\subsection{Consumer Client}

The Consumer Client, shown in region 3 of Figure~\ref{fig:memory-architecture}, connects a Coding Assistant through MCP to the shared Knowledge Base while the assistant continues to manage its Local Knowledge Base. Human Users access the same store through a UI, and the backing service also exposes REST APIs. Retrieval supports three tiers: Fast uses tag-filtered hybrid keyword/vector search, Normal expands and reranks the candidate pool, and Deep Research iteratively follows linked memories and synthesizes a cited answer.

Rather than raw database rows, the client receives structured memories containing the question, answer, examples, caveats, tags, trust score, and related-memory links. These links let the Coding Assistant traverse the memory graph through MCP by fetching related memories as needed.

Authenticated MCP and UI feedback is limited to one vote per identity, memory, and agent/human channel; accepted votes update counters and recalculate a 0-10 heuristic trust score from examples, caveats, recency, and Bayesian feedback. The feedback term is $B=(vR+mC)/(v+m)$, where $v$ is the vote count, $R$ the observed upvote ratio, $C$ the prior mean, and $m$ the prior weight. Shrinking sparse feedback toward the prior prevents a few early votes from producing extreme scores while allowing sustained evidence to dominate. The score has not been calibrated against downstream task success.

\section{Design Considerations}

Enterprise coding memory must align stored knowledge with later queries, yet raw traces, diffs, and implementation notes are often poorly aligned. \citet{ma2023queryrewriting} describe this retrieval problem, observing that ``there is inevitably a gap between the input text and the needed knowledge in retrieval.'' We therefore treat memory format as part of retrieval design and store Curated Memory as question--answer units containing a developer-facing question, direct answer, working examples, and caveats. The Consumer Client searches both fields: question matching serves problem-shaped queries, while answer matching captures implementation details, error messages, and API names.

Shared memory also needs governance because one corrupted memory can influence many future agents. Recent work on sleeper memory poisoning shows that ``a single exposure may corrupt future behavior even after the original malicious context is no longer visible'' \citep{pulipaka2026hidden}. The system therefore uses gating, curation, trust scoring, and feedback to prevent unsafe or low-quality memories from becoming organization-wide guidance, and to keep accepted memories revisable rather than permanently authoritative.

\section{Deployment Snapshot}

At the July 22, 2026 snapshot, the deployed pipeline had processed \textbf{900 contributed learnings}: \textbf{483 (53.7\%) yielded at least one persisted memory} and \textbf{417 (46.3\%) yielded none}, producing \textbf{1,144 curated memories}, or 2.37 per productive learning. This snapshot is limited to selected internal teams at SAP SE and should not be interpreted as a company-wide deployment. Although current telemetry cannot identify the responsible pipeline stage, preliminary developer reviews indicated that the gates were operating as intended, with almost all reviewed no-yield cases reflecting repeated knowledge; finer-grained stage-level disposition logging is planned. Of the persisted memories, \textbf{1,100 (96.2\%)} had at least one semantic link, with \textbf{7,863 links} across \textbf{35 tools and frameworks}. These figures characterize deployment scale and throughput, not retrieval relevance or downstream utility.

\section{Conclusion}

Existing shared-memory systems make agent learning reusable but leave contribution dependent on agents recognizing and explicitly recording useful lessons, disconnecting knowledge capture from the coding workflow. Our ongoing production deployment addresses this gap with a three-stage organizational memory substrate: the Contributor Client initiates task-adjacent capture with user approval, the Curation Pipeline converts accepted contributions into governed, trust-aware question--answer memories, and the Consumer Client retrieves them through MCP when future agents need them. Together, these deployed components turn private development experience into a managed, reusable knowledge base; ongoing work evaluates retrieval relevance, downstream coding-task utility, memory freshness, and calibration of the heuristic trust score.

\section{Limitations}

Our system does not make memory capture completely invisible. To avoid tracking an entire codebase, collection must be scoped to designated memory files, and the coding assistant must be instructed to write reusable learnings there. In practice, this requires assistant-facing configuration, such as \texttt{AGENTS.md} or assistant-specific configuration directories such as \texttt{.claude}, to make memory writing part of the working environment. This setup step is important: it keeps the capture surface small and reviewable, but it also means the system only observes learnings that are externalized into the configured memory surface.

Feedback has a similar dependency. The system can serve memories during coding tasks and record agent or human feedback, but feedback still requires a tangential action after the memory is consumed. The same assistant-facing configuration must encourage agents to upvote useful memories and flag incorrect ones, especially when retrieved guidance is wrong. A more autonomous design would use a separate background memory process to consolidate usage history and update memory state, similar to recent dreaming approaches in OpenAI Memory Dreaming and Anthropic Dreams \citep{openai-dreaming,anthropic-dreams}. Such architectures are not yet a universal deployment assumption for enterprise coding assistants.

The curation pipeline is also limited by its distance from the original task. It curates a memory after the fact: it did not solve the problem, observe every constraint, or validate that the resulting guidance will work in future contexts. The system therefore treats curation primarily as a conservative rewrite step, transforming captured development evidence into a retrievable question--answer memory rather than adding new substantive claims. Even so, a memory that is incorrect but not suspicious may pass through curation and be served before enough negative feedback accumulates to lower its trust score. During that window, consuming agents may receive unhelpful guidance, which makes supervision and feedback an important part of the architecture.

The security-gate sanity check has a related circularity limitation: the system developers authored its synthetic cases, acceptable-action sets, and deterministic postconditions without independent adjudication, so the reported agreement may reflect the gate's design assumptions.

Finally, the system reduces but cannot eliminate memory-based injection risk. Curation, memory enrichment, trust scoring, and gated persistence can filter low-quality or suspicious memories before they are served, but a shared memory system remains part of the agent's input surface. The final safeguard is still the supervised coding workflow in which retrieved guidance is inspected, applied, tested, and reviewed. For this reason, it is intended as an internal organizational knowledge base with authenticated read and write access, traceable contribution provenance, and curation before persistence, rather than as an unauthenticated public memory store.

\section{Ethical Considerations}

The system processes private development knowledge that may contain proprietary code, credentials, personal data, or security-sensitive infrastructure details. Restricting collection to designated memory files, performing contextualization locally, and requiring user approval reduce unnecessary disclosure but cannot eliminate it. The deployed gate applies secret and personal-data scanning; production hardening additionally requires encryption, tenant isolation, least-privilege access, auditable access controls, and retention and deletion policies.

A shared memory store can amplify incorrect, stale, or malicious guidance. Curation, provenance, validation gates, trust scores, and feedback reduce but do not eliminate memory poisoning. Retrieved memories should therefore remain advisory and subject to inspection, testing, security review, and human approval for consequential changes.

Authentication is needed to verify authorization and provide accountability for abuse or security incidents, but it must not enable productivity surveillance. Routine records, curation interfaces, and aggregate reports should use pseudonymous contributor identifiers, with identity mappings stored separately under strict access control and disclosed only for documented security or compliance investigations. Such access should be purpose-limited, time-limited, and logged. Contribution counts, timestamps, and feedback histories must not be repurposed for performance assessment. Contributors should be informed about collection and retention and, where policy permits, be able to correct or remove their memories.

\paragraph{AI Use Statement.} Generative AI was used as a writing assistant in preparing this paper. It supported drafting, editing, and language refinement, and was used to generate Figure~\ref{fig:automatic-contribution-flow} from author-specified content and design constraints. The authors reviewed and verified the resulting text, checked the figure's technical content, manually cropped the figure for publication, and take responsibility for the paper's content.

\bibliographystyle{plainnat}
\bibliography{references}

\clearpage
\appendix

This appendix retains the implementation and evaluation details needed to
interpret the paper's claims without reproducing full source schemas, prompts,
fixtures, or the complete synthetic case manifest. Deployment-specific
organization and infrastructure identifiers are replaced by neutral
descriptions; model identities and reported tunable values are retained.

\section{Captured Learning Schema}
\label{app:captured}

The Contributor Client sends each approved edit to the Unprocessed Knowledge Store as a raw learning. Before enrichment, the Curation Pipeline retains the textual patch for provenance and renders it into structured hunks. Identifiers and source groups are pseudonymous. Table~\ref{tab:raw-learning} summarizes the collection contract.

\begin{table}[htbp]
\centering
\begin{tabular}{@{}p{0.30\columnwidth}p{0.60\columnwidth}@{}}
\hline
\textbf{Field} & \textbf{Description} \\
\hline
\texttt{learning\_id} & Contribution UUID (provenance). \\
\texttt{user\_id} & Contributor pseudonym. \\
\texttt{team\_name} & Pseudonymous source-group identifier. \\
\texttt{learning\_type} & Memory-file / path category. \\
\texttt{patch} & Diff Match Patch textual edit. \\
\texttt{sent\_at} & Contribution timestamp. \\
\hline
\end{tabular}
\caption{Raw learning fields at collection intake.}
\label{tab:raw-learning}
\end{table}

Patch rendering expands \texttt{patch} into one or more \emph{hunks}. Each hunk records \texttt{change\_number}, lowercase \texttt{change\_type} (\texttt{replace}, \texttt{insert}, or \texttt{delete}), removed and added lines, up to five context lines on each side, and the nearest headings with optional line numbers. The validated model requires \texttt{total\_changes} to equal the hunk count; the deployed intake path does not yet enforce this invariant.

\section{Curated Memory Schema}
\label{app:curated}

Every persisted memory conforms to a validated structured contract.
Table~\ref{tab:curated-memory} marks whether a field is
\textbf{S}earchable, \textbf{D}isplayed to the consuming agent, or used only
for \textbf{G}overnance and provenance.

\begin{table}[htbp]
\centering
\begin{tabular}{@{}p{0.30\columnwidth}cp{0.44\columnwidth}@{}}
\hline
\textbf{Field} & \textbf{Role} & \textbf{Description} \\
\hline
\texttt{id} & G & Memory UUID. \\
\texttt{source\_id} & G & Origin learning id. \\
\texttt{user\_id} & G & Contributor pseudonym. \\
\texttt{team\_name} & G & Pseudonymous source group. \\
\texttt{question} & S,D & Developer-facing question. \\
\texttt{answer} & S,D & Direct answer. \\
\texttt{working\_example} & D & Concrete code/scenario. \\
\texttt{caveats} & D & Edge cases, conditions. \\
\texttt{linked\_questions} & D & Related-memory links. \\
\texttt{tool\_tags} & S,D & Tool/platform tags. \\
\texttt{type\_tags} & S,D & Knowledge-type tags. \\
\texttt{lob} & G & Line of business. \\
\texttt{trust\_score} & D & 0--10 reliability signal. \\
\texttt{*\_upvotes} / & G & Agent and human feedback \\
\texttt{*\_downvotes} & & counters. \\
\texttt{created\_at} & G & Creation time (recency). \\
\texttt{security\_scan} & G & Optional gate-audit metadata. \\
\hline
\end{tabular}
\caption{Curated memory fields and their retrieval / governance roles.}
\label{tab:curated-memory}
\end{table}

Question and answer are embedded separately, tags support filtering, and governance-only fields are not searchable. The optional \texttt{security\_scan} property is reserved for gate-audit metadata. The production persistence path consumes the decision without storing this metadata.

\section{Curation Agent Contracts}
\label{app:agents}

The pipeline uses four LLM-agent roles. The rubrics below summarize the
behavioral contracts used in the reported deployment.

\paragraph{Execution configuration.} Every LLM call used Claude Sonnet 4.5 with \texttt{temperature=0}. \texttt{max\_tokens} was unset except for the tag-validation call (150). Target JSON Schemas were injected into system prompts and outputs validated with Pydantic; responses were parsed from raw JSON. Transient connection, timeout, and 5xx errors received up to three retries with exponential backoff (1\,s initial, $\times 2$, 30\,s cap).

\paragraph{Enrichment Agent.} In a first call, filter hunks with insufficient context; in a second, convert surviving hunks into zero or more self-contained question--answer memories with a developer-facing question, direct answer, copied working example, and caveats when present. Merge related hunks, split dense hunks, and emit nothing when no reusable knowledge exists. Do not introduce claims unsupported by the captured evidence.

\paragraph{Bucketing Agent.} First assign \texttt{tool\_tags} and \texttt{type\_tags} from the shared registry, then validate any proposed new tag in a separate call. Reuse an existing tag whenever one fits; approve a new tag only when it is distinct, reusable, and clearly named. Deterministic tag identifiers make upserts idempotent. If validation leaves a dimension empty, use the generic fallbacks \emph{General} and \emph{Best Practice}.

\paragraph{Gating (Security Scan) Agent.} Treat each candidate in the deployed curation pipeline as a lightweight, non-exhaustive scan target (Appendix~\ref{app:gate}); call at most six relevant scanner tools in parallel; disclose only the content each check needs; emit per-category status, optional rewrite, and comments; and preserve useful grounded content when rewriting.

\paragraph{Collision Checker.} Retrieve existing memories by tag-filtered keyword and vector similarity over both question and answer. The first call passes candidates with no close match or accepts, clarifies, or rejects them as duplicate or low value; a separate linking call optionally connects retained memories to prerequisites, follow-ups, or alternatives.

\section{Security and Privacy Gate}
\label{app:gate}

This section specifies the gate deployed in the production curation pipeline. It is placed after enrichment and before trust scoring, bucketing, collision checking, and persistence. Each candidate is scored across six categories using \emph{Safe}, \emph{Unsafe}, \emph{Safe with caveats}, or \emph{N.A}. For fixed input and configuration, its in-repository rule-based checks are reproducible; they are not pinned invocations of external scanner binaries, and agent routing and rewriting can vary unless the model, prompt, and decoding settings are pinned. We evaluate its decision behavior through the offline functional sanity check in Appendix~\ref{app:gate-eval}.

\begin{table}[htbp]
\centering
\begin{tabular}{@{}p{0.34\columnwidth}p{0.56\columnwidth}@{}}
\hline
\textbf{Category} & \textbf{In-repository check} \\
\hline
Dependency & OSV/GuardDog-style checks \\
Secret & Gitleaks-style check \\
Code & Bandit/Semgrep-style checks \\
PII / internal & Presidio-style check \\
Workflow & Command-policy scanner \\
Grounding & Evidence consistency check \\
\hline
\end{tabular}
\caption{Gate categories and corresponding in-repository checks.}
\label{tab:gate-scanners}
\end{table}

Category statuses aggregate by severity: any \emph{Unsafe} dominates, then \emph{Safe with caveats}, \emph{Safe}, and \emph{N.A}. The aggregate status drives one persistence decision (Table~\ref{tab:gate-decision}); there is no quarantine state. Secret and PII checks can return replacement hints. Persisting statuses, action, and comments in a governance-only, non-searchable \texttt{security\_scan} field remains a design property: the production path consumes the decision and discards it. The sanity check below measures behavior on synthetic cases rather than the effectiveness of pinned external scanner versions.

\begin{table}[htbp]
\centering
\begin{tabular}{@{}p{0.40\columnwidth}p{0.50\columnwidth}@{}}
\hline
\textbf{Aggregate outcome} & \textbf{Pipeline action} \\
\hline
Any category Unsafe & Reject (drop); block persistence. \\
Safe with caveats, caveat added to rewrite & Persist rewritten memory. \\
Safe with caveats, caveat not expressible & Escalate to Unsafe; block. \\
All Safe or N.A & Persist candidate unchanged. \\
\hline
\end{tabular}
\caption{Gate decision table mapping aggregate status to action.}
\label{tab:gate-decision}
\end{table}

\subsection{Offline Functional Sanity Check}
\label{app:gate-eval}

\paragraph{Objective and scope.}
We performed a controlled offline evaluation to determine whether the deployed Security Scan Agent exhibits its intended basic behavior on purpose-built synthetic inputs. The evaluation is a functional sanity check rather than a security benchmark: its manually specified cases are not representative of production memories and do not establish robustness to adaptive attacks, production data, or the full range of possible security defects. The evaluated implementation is the gate used in the production curation pipeline.

Each execution invoked the same \texttt{EnrichmentResult}-to-\texttt{Memory} transformation and \texttt{GatingAgent.process()} path as production. Evidence was supplied as a serialized synthetic \texttt{RenderedPatch}; before invocation, the harness validated learning and contributor identifiers, \texttt{source\_changes}, and any verbatim \texttt{working\_example} against that evidence. Postconditions were deterministic: exact category or action where required, identity preservation across rewrites, and required or forbidden content. No human or LLM judge was used. For secret and personal/internal-data cases, forbidden-content checks verified that every seeded synthetic sensitive value was absent from an accepted rewrite.

\paragraph{Test suite.}
The suite contained 56 cases: 33 memories seeded with a security or privacy risk and 23 benign controls. Table~\ref{tab:gate-eval-suite} gives the distribution. The suite exercises five security-relevant categories: dependencies, secrets, unsafe code, personal or internal information, and unsafe workflows. Although the decision contract also emits a grounding status, the suite did not contain a separate adversarial grounding evaluation.

\begin{table}[htbp]
\centering
\begin{tabular}{@{}lrrr@{}}
\hline
\textbf{Case group} & \textbf{Risk} & \textbf{Benign} & \textbf{Total} \\
\hline
Dependency & 4 & 1 & 5 \\
Secret & 4 & 1 & 5 \\
Code & 16 & 0 & 16 \\
PII / internal & 4 & 1 & 5 \\
Workflow & 5 & 0 & 5 \\
Additional safe controls & 0 & 20 & 20 \\
\hline
Total & 33 & 23 & 56 \\
\hline
\end{tabular}
\caption{Composition of the synthetic security-gate sanity-check suite.}
\label{tab:gate-eval-suite}
\end{table}

\paragraph{Representative paired cases.}
Table~\ref{tab:gate-representative-cases} illustrates the distinction tested
between risky constructs and superficially similar safe controls. The complete
suite also covered vulnerable or suspicious dependencies, additional secret
formats, unsafe deserialization, disabled TLS verification, path traversal,
server-side request forgery, archive extraction, JWT verification, AES-GCM
nonce reuse, regular-expression denial of service, internal identifiers, and
unsafe operational commands.

\begin{table}[htbp]
\centering
\small
\begin{tabular}{@{}p{0.15\columnwidth}p{0.36\columnwidth}p{0.36\columnwidth}@{}}
\hline
\textbf{Category} & \textbf{Seeded-risk example} & \textbf{Matched safe control} \\
\hline
Secret & Synthetic API key in useful guidance & Documented token placeholder \\
Command & User input interpolated in a shell command & Subprocess argument list without a shell \\
Database & SQL assembled by string interpolation & Parameterized SQL query \\
Web & Untrusted content via \texttt{innerHTML} & Safe content via \texttt{textContent} \\
Template & Autoescaping disabled for untrusted text & Ordinary autoescaped template \\
Memory safety & Unbounded copy into a fixed buffer & Copy bounded by destination capacity \\
\hline
\end{tabular}
\caption{Representative risk-bearing cases and matched benign controls.}
\label{tab:gate-representative-cases}
\end{table}

\paragraph{Expected-action semantics.}
The expected labels specify acceptable persistence actions, not interchangeable classifier labels. \emph{Blocked} rejects the candidate; \emph{Rewritten} persists a replacement only after the seeded risk is removed or replaced; \emph{Caveated} persists a replacement carrying an explicit safety qualification; and \emph{Unchanged} persists a benign input as supplied. A case lists two or three outcomes when more than one resolution is safe: for example, a dangerous command may be blocked or removed in a rewrite, while uncertain dependency guidance can be rewritten or retained with a required caveat. Benign controls marked \emph{Caveated/Unchanged} permit an optional non-blocking warning. In every case, an action is accepted only when deterministic assertions also preserve identity and required guidance and exclude seeded secrets, personal or internal values, or unsafe patterns as applicable.

\paragraph{Repeated executions.}
Because an LLM coordinates the category decisions and optional rewriting, we executed the complete 56-case suite three times with identical inputs and model configuration, using Claude Sonnet 4.5. This produced 168 decisions. The repeated runs measure outcome consistency for one configured model; they do not establish portability across models or providers. The runner requires model credentials but no database, vector store, or API server.

\paragraph{Practical outcome categories.}
We classified each execution according to its effect on memory persistence, rather than relying only on the test harness's internal pass/fail labels:

\begin{itemize}
    \item \textbf{Appropriate:} a seeded risk was blocked, safely rewritten, or handled with an allowed caveat, or a benign memory passed unchanged.
    \item \textbf{Cautious, non-blocking:} the gate behaved more conservatively than required but preserved the memory. This includes adding a caveat to benign content and returning a caveated replacement where the stricter expected action was blocking or rewriting. In such cases, the risky element was removed or explicitly qualified.
    \item \textbf{Overcautious, blocking:} the gate rejected a memory that was expected to remain persistable.
    \item \textbf{Unsafe acceptance:} unsafe content remained persistable without adequate removal, rewriting, or explicit qualification.
\end{itemize}

\paragraph{Results.}
Table~\ref{tab:gate-eval-results} reports the practical outcomes. Across all 168 executions, 135 (80.4\%) exhibited the expected behavior. A further 32 (19.0\%) were cautious but non-blocking: they preserved the memory while adding a warning or applying a more conservative transformation. One execution (0.6\%) was overcautious and blocked an expected-safe memory. No execution was classified as an unsafe acceptance.

\begin{table}[htbp]
\centering
\begin{tabular}{@{}lrrrr@{}}
\hline
\textbf{Practical outcome} & \textbf{R1} & \textbf{R2} & \textbf{R3} & \textbf{Total} \\
\hline
Appropriate & 46 & 45 & 44 & 135 (80.4\%) \\
Cautious, non-blocking & 9 & 11 & 12 & 32 (19.0\%) \\
Overcautious, blocking & 1 & 0 & 0 & 1 (0.6\%) \\
Unsafe acceptance & 0 & 0 & 0 & 0 (0.0\%) \\
\hline
Total & 56 & 56 & 56 & 168 \\
\hline
\end{tabular}
\caption{Practical outcomes across three complete executions of the 56-case suite.}
\label{tab:gate-eval-results}
\end{table}

The 33 risk-bearing cases produced 99 executions. Of these, 87 were classified as appropriate and 12 as cautious but non-blocking; none produced an unqualified unsafe acceptance. The 23 benign cases produced 69 executions: 48 passed as expected, 20 remained persistable with additional caveats, and one was blocked. Thus, 68 of 69 benign executions (98.6\%) remained persistable, although 20 of them received unnecessary or optional cautionary text.

\paragraph{Consistency across runs.}
The exact persistence action---blocked, rewritten, caveated, or unchanged---was identical across all three runs for 37 of 56 cases (66.1\%). When actions that have the same practical safety effect are grouped using the four categories above, 43 of 56 cases (76.8\%) retained the same classification across all three runs.

Thirteen cases changed practical category between runs. Twelve alternated only between appropriate behavior and cautious non-blocking behavior. The remaining case, the safe resolved-path containment control, was blocked in the first run but retained with caveats in the other two. No case changed from an appropriate or cautious outcome to unsafe acceptance. These results indicate that the LLM affects whether the gate blocks, rewrites, or adds caveats, while the observed variation was predominantly conservative rather than permissive.

\paragraph{Interpretation.}
The experiment supports the limited claim that the gate performs its intended basic filtering function on this controlled suite: seeded risks consistently triggered blocking, rewriting, or explicit qualification, and nearly all benign inputs remained persistable. The relatively frequent cautious outcomes also show that the gate tends to add warnings to safe or remediated content. The single overcautious block and the variation between blocking, rewriting, and caveating motivate continued calibration and monitoring in production. Because the cases are synthetic, use one configured LLM, and do not include adaptive attacks or production traffic, the results should not be interpreted as a general security guarantee.

\section{Retrieval Interfaces}
\label{app:retrieval}

The serving layer exposes the shared Knowledge Base through REST and MCP. MCP provides \texttt{query} for Fast or Normal retrieval, \texttt{deep\_research} for iterative synthesis, \texttt{read\_memory} for one memory and resolved link triples, and \texttt{feedback} for up- or down-votes. Table~\ref{tab:retrieval-config} records the implemented retrieval defaults.

\begin{table}[htbp]
\centering
\begin{tabular}{@{}p{0.30\columnwidth}p{0.60\columnwidth}@{}}
\hline
\textbf{Setting} & \textbf{Implemented value} \\
\hline
Fast & Fetch and return \texttt{top\_x} (default 5; maximum 50); no reranking. \\
Normal & Fetch $\max(20,2\,\texttt{top\_x})$, rerank with \texttt{cohere-reranker}, return \texttt{top\_x}. \\
Hybrid alpha & 0.5 (equal BM25/vector blend). \\
Q/A fusion & Question 0.7; answer 0.3. \\
Collision pool & Top 5 after an overlapping-tag prefilter. \\
Min. similarity & 0.3 declared; not yet enforced. \\
Explicit tags & AND over \texttt{tool\_tags}. \\
Automatic tags & Tool-dimension matching only. \\
Reranker failure & Preserve the pre-rerank order. \\
\hline
\end{tabular}
\caption{Implemented Fast and Normal retrieval configuration.}
\label{tab:retrieval-config}
\end{table}

Fast performs tag matching, query embedding, and hybrid multi-vector search. Normal expands the pool and reranks it; Deep Research iteratively searches questions and answers, follows links, and synthesizes a cited response. The REST interface accepts \texttt{question}, \texttt{mode}, optional \texttt{tags}, and \texttt{top\_x}. Query results return links compactly; \texttt{read\_memory} resolves them for graph traversal.

The Weaviate \texttt{Memories} collection maintains separate, self-provided question and answer vectors, each with an HNSW index using the database-default cosine distance, alongside a BM25 inverted index. Both curation and serving use \texttt{gemini-embedding} with 1536 output dimensions (from the model's native 3072), so all question and answer vectors share one embedding configuration.

\section{Trust Score Computation}
\label{app:trust}

The trust score is a 0--10 signal with five components: a working example (2 points), caveats (1 point), and recency (1 point) are deterministic; agent feedback (4 points) and human feedback (2 points) are Bayesian. Recency awards the full point for the first 30 days and decays linearly to zero at two years (730 days).

Each feedback component scales its maximum by the Bayesian average
\[
B=\frac{vR+mC}{v+m},
\]
where $v=\text{upvotes}+\text{downvotes}$, $R=\text{upvotes}/v$ is the observed ratio, $C=0.5$ is the prior mean, and $m$ is the prior weight ($m=10$ for agent votes, $m=3$ for human votes). With no votes ($v=0$) the component returns $C$ times its maximum, holding the score near the prior.

\paragraph{Resistance to a few early votes.} A memory with a working example ($+2$), caveats ($+1$), and age under 30 days ($+1$) starts with 4 deterministic points and scores $7.0$ before receiving feedback. Suppose an attempt is made to inflate it with only 3 agent upvotes and 1 human upvote, all positive:
\begin{align*}
B_a &= \frac{3+10(0.5)}{3+10}=0.615, &4B_a&=2.46,\\
B_h &= \frac{1+3(0.5)}{1+3}=0.625, &2B_h&=1.25.
\end{align*}
The total becomes only $4+2.46+1.25=7.71$: four unanimous votes move the score by $0.71$, rather than driving it near 10, illustrating the prior's damping effect. The authenticated serving path immediately recalculates trust and limits feedback to one vote per identity and memory in each agent or human channel. This prevents repeated-vote inflation by one account; coordinated abuse across authenticated identities still requires monitoring.

\end{document}